\documentclass[authoryear]{elsarticle}
\usepackage{lineno}
\usepackage[utf8]{inputenc}
\usepackage{url}
\usepackage{hyperref}
\usepackage{amsfonts}
\usepackage{algorithm2e}
\usepackage{amsmath}
\usepackage{subcaption}
\usepackage{booktabs}
\usepackage{lmodern}
\usepackage{microtype}      
\usepackage{nicefrac}       
\usepackage{setspace}
\usepackage[flushleft]{threeparttable}  
\usepackage{siunitx}
\usepackage{comment}  
\newcolumntype{H}{>{\setbox0=\hbox\bgroup}c<{\egroup}@{}}  

\hypersetup{pdfauthor=author} 

\newcommand{\fscore}{\text{F}_1\text{-score}}

\newcommand{\tons}[1]{#1\si{\tonne}}

\usepackage[color=orange!50]{todonotes}
\usepackage{soul}

\definecolor{ballblue}{rgb}{0.13, 0.67, 0.8}
\definecolor{sapgreen}{rgb}{0.31, 0.49, 0.16}
\definecolor{candypink}{rgb}{0.89, 0.44, 0.48}
\definecolor{ao(english)}{rgb}{0.0, 0.5, 0.0}

\begin{document}

\begin{frontmatter}
\title{How do tuna schools associate to dFADs? A study using echo-sounder buoys to identify global patterns}

\author[1,2]{Manuel Navarro-Garc\'ia}
\ead{mannavar@est-econ.uc3m.es}

\author[3]{Daniel Precioso}
\ead{daniel.precioso@uca.es}

\author[5]{Kathryn Gavira-O'Neill}
\ead{kgo@satlink.es}

\author[2]{Alberto Torres-Barr\'an}
\ead{alberto.torres@komorebi.ai}

\author[2]{David Gordo}
\ead{david.gordo@komorebi.ai}

\author[2]{V\'ictor Gallego}
\ead{victor.gallego@komorebi.ai}

\author[3,4]{David G\'omez-Ullate\corref{corauthor}}
\ead{david.gomezullate@uca.es}

\cortext[corauthor]{Corresponding author}
\address[1]{Universidad Carlos III de Madrid, Calle Madrid, 126, 28093 Getafe, Madrid, Spain.}
\address[2]{Komorebi AI Technologies, Avenida General Per\'on Nº 26, Planta 4ª,
28020 Madrid, Spain.}
\address[3]{Universidad de C\'adiz,  Av. Universidad de C\'adiz, 10, 11519 Puerto Real, C\'adiz, Spain.}
\address[4]{School of Science and Technology, IE University, Segovia, Spain.}
\address[5]{Satlink S.L.U., Carretera de Fuencarral, Arbea Campus Empresarial Edificio 5. Planta baja, 28108 Alcobendas, Madrid, Spain.}

\begin{abstract}
Based on the data gathered by echo-sounder buoys attached to drifting Fish Aggregating Devices (dFADs) across tropical oceans, the current study applies a Machine Learning protocol to examine the temporal trends of tuna schools' association to drifting objects. Using a binary output, metrics typically used in the literature were adapted to account for the fact that the entire tuna aggregation under the dFAD was considered. The median time it took tuna to colonize the dFADs for the first time varied between 25 and 43 days, depending on the ocean, and the longest soak and colonization times were registered in the Pacific Ocean. The tuna schools' Continuous Residence Times were generally shorter than Continuous Absence Times (median values between 5 and 7 days, and 9 and 11 days, respectively), in line with the results found by previous studies. Using a regression output, two novel metrics, namely aggregation time and disaggregation time, were estimated to obtain further insight into the symmetry of the aggregation process. Across all oceans, the time it took for the tuna aggregation to depart from the dFADs was not significantly longer than the time it took for the aggregation to form. The value of these results in the context of the ``ecological trap" hypothesis is discussed, and further analyses to enrich and make use of this data source are proposed.

\end{abstract}

\begin{keyword}
Tropical tunas \sep Echo-sounder buoys \sep Fish Aggregating Devices (FADs) \sep Associative behaviour \sep Ecological trap
\end{keyword}

\end{frontmatter}
\section{Introduction} 

For centuries, floating objects drifting on the ocean’s surface have been known to attract a number of fish species, including tropical tunas such as skipjack tuna (\textit{Katsuwonus pelamis}), yellowfin tuna (\textit{Thunnus albacares}) and bigeye tuna (\textit{Thunnus obesus}) \citep{Castro2002AHypothesis,Maufroy2015Large-ScaleOceans}. As fishermen have noticed this behaviour, they have used both natural and man-made floating objects, or drifting Fish Aggregating Devices (dFADs), as a tool for finding and catching tropical tunas. The use of dFADs in tuna purse-seine fisheries has gradually increased since the 1980s to the present time, where vessels using dFADs now contribute to $36\%$ of the world’s total tropical tuna catch ~\citep{Davies2014TheOcean,Wain2021QuantifyingFisheries,ISSF2021Status2021}. These widespread changes have highlighted the need to better understand the potential ecological effects of dFADs on tuna ecology and the marine environment, in order to ensure adequate management of fish stocks and dFAD usage. 

Indeed, both the dynamics of how and why tuna associate to dFADs are still poorly understood. Regarding the reasons behind tuna aggregation to dFADs, a number of hypotheses have been suggested \citep{Freon2000ReviewHypothesis,Dempster2004FishStudies, Castro2002AHypothesis}. Of these, two have gained traction: the ``meeting-point" hypothesis, which considers that dFADs facilitate the encounter between individuals or schools, thus constituting larger schools that could benefit survival rates \citep{Castro2002AHypothesis}; and the ``indicator-log" hypothesis, by which tunas may be safeguarding the survival of their eggs, larvae and juvenile stages by using drifting objects as indicators of areas where plankton and food is readily available \citep{Hall1992CharacteristicsTunas}. This scenario has led some authors to postulate that man-made dFADs could have detrimental effects on tuna populations by creating a so-called ``ecological trap" which would lead tuna to remain associated to dFADs even as these drift into areas that could negatively affect the tuna's behaviour and biology \citep{Marsac2000DriftingTrap, Hallier2008DriftingSpecies}. To the best of our knowledge, there is yet no sufficient evidence to either confirm or reject this hypothesis (see \citet{dagorn2012IsEcosystems} and references therein).  

Given the concerns around the widespread use of dFADs in tuna fisheries today, it is not surprising that a considerable amount of  research has been devoted to characterizing the dynamics at play when tunas aggregate to dFADs. However, results appear to be highly variable. The \textit{continuous residence time} of tunas at dFADs, defined as the duration for which tuna was present at the FAD without day-scale absences \citep{Ohta2004PeriodicStations}, has been found in the literature to range from less than a day to 55 days \citep{Baidai2020TunaBuoys}. Likewise, the values of \textit{continuous absence time}, i.e., the time between two consecutive associations to dFADs \citep{Robert2012Size-dependentFADs}, ranges from 2 days to over 100 days \citep{Robert2012Size-dependentFADs, Baidai2020TunaBuoys}. Given the inherent difficulties of conducting experiments in the open ocean, most research on this subject is based on small-scale studies using electronic or acoustic tags to monitor individual tunas at a small number of dFADs, which might explain the variability in these results.

However, the dFADs used by tropical tuna purse-seine fisheries today are generally deployed with satellite-linked instrumented buoys equipped with one or more echo-sounders, which provide fishermen accurate dFAD positioning as well as estimates of aggregated tuna biomass \citep{Davies2014TheOcean, Wain2021QuantifyingFisheries}. Data collected by these buoys provide invaluable information for fishermen, but have also attracted the attention of the scientific community, who have highlighted their potential to provide insights in tuna migration and behaviour on a global scale \citep{Santiago2016TowardsTT-BAI,Baidai2020MachineData,Orue2019FromBuoys,Lopez2016ADevices}. As such, recent studies have begun to model and process the echo-sounder data provided by these buoys to remotely map tuna distribution, or investigate patterns in tuna aggregation around dFADs \citep{Baidai2020TunaBuoys, Baidai2019MappingFrom, Precioso2022Tun-AI:Data, Orue2019AggregationSpecies}. 

In this context, the current study applies the Machine Learning based models from \textsc{Tun-AI} \citep{Precioso2022Tun-AI:Data} to provide accurate biomass estimates below dFADs across the Atlantic, Indian and Pacific Oceans, with the aim of characterizing the temporal patterns of tuna associations to dFADs. To do this, we adapt metrics already present in the literature to account for the fact that our study focuses on the entire tuna aggregation around the dFAD, as opposed to individual fish. Given that \textsc{Tun-AI} can deliver estimated amounts of tuna biomass aggregated to the dFAD, we examine the processes of aggregation and disaggregation in more detail. We check whether there could be a potential ``ecological trap'' \citep{Marsac2000DriftingTrap, Hallier2008DriftingSpecies} effect on the tuna schools, by testing whether the time it takes for the tuna school to depart from the dFAD is significantly longer than the time it takes for the aggregation to form in the first place.

\section{Material and methods}\label{sec: material and methods}

\subsection{Database description}

The work presented in this paper makes use of an extensive amount of data arranged in three large databases, classified according to the source they were obtained from. 

\subsubsection{Activity data on dFADs} \label{subsubsec: activity data}

The first database contains the activities performed by the Spanish tropical tuna purse seine fleet on dFADs drifting in the three major oceans (Atlantic, Indian and Pacific). These data were provided by the ship owner's association, Asociación de Grandes Atuneros Congeladores (AGAC), and contains \num{120707} events spanning between 11th April 2017 and 1st January 2021, out of which \num{35813} happened in the Atlantic Ocean, \num{55819} in the Indian Ocean, and the remaining \num{29075} in the Pacific Ocean. Every entry in this database contains information on the type of interaction with the dFAD, the unique identification number and the model of the echo-sounder buoy attached to the dFAD, the timestamp and geographical coordinates where the activity took place, and other relevant details (for a complete description of the interaction types, see \citet{Ramos2017SpanishRequirements}). The buoy identification number allowed us to establish a connection between the human interactions associated to this dFAD and the acoustic measurements recorded by the echo-sounder (see Section~\ref{subsubsec: Echo-sounder buoy data}).


\subsubsection{Echo-sounder buoy data} \label{subsubsec: Echo-sounder buoy data}

The echo-sounder buoy database assembles the data collected from Satlink (\url{www.satlink.es}) buoys  deployed by the Spanish tropical tuna purse seine fleet. Altogether, this data set includes information from \num{16419} different buoys distributed over the three major oceans and spanning  the same time frame as the events in the activity database (see Section~\ref{subsubsec: activity data}). The data set comprises over 70 million  observations, generally sampled at hourly frequency. 

Each entry in the database contains the unique buoy identification number, the timestamp of when the reading was taken, and an estimate of tuna biomass under the dFAD. These biomass estimates (in metric tons, \tons{}) are obtained from acoustic samples taken periodically throughout the day, and the average back-scattered acoustic response is converted into estimated tonnage, based on the target strength of skipjack tuna (see~\citet{Lopez2016ADevices} for detailed explanations of the process within the buoy). For each reading, the biomass estimates are provided across ten equally spaced depth layers, and values can range from \tons{0} to \tons{63}. 

The data set also includes all the position information transmitted by the buoy. These GPS coordinates of the buoy are generally transmitted every 24 hours, although transmission frequency can be modified by the buoy owner. Besides that, buoys are programmed to only send biomass estimates when the total measurement delivers values above \tons{1}. Hence, if a given buoy sent GPS coordinates but no biomass estimates over a certain period, the biomass estimates for that period were imputed to \tons{0}. Further information about the buoy models and the biomass estimation process is available in \citet{Precioso2022Tun-AI:Data}, Section~2.1.2.

\subsubsection{Oceanography data} \label{subsubsec: oceanography}
The \textsc{Tun-AI} models are trained to provide accurate biomass estimates from echo-sounder data \citep{Precioso2022Tun-AI:Data}, but they also need to be fed with several oceanographic variables at surface level (depth~$=0.494\text{m}$). These data are provided at daily frequency by the EU Copernicus Marine Environment Monitoring Service (CMEMS)~\citep{GlobalMonitoringandForecastingCenter2018OperationalInformation} (products GLOBAL-ANALYSIS-FORECAST-PHY-001-024, $1/12^\circ$ resolution; and GLOBAL-ANALYSIS-FORECAST-BIO-001-028, $1/4^\circ$ resolution). Each record of the echo-sounder buoy database (see Section~\ref{subsubsec: Echo-sounder buoy data}) is enriched with oceanographic variables for the location and time of the measurement.

\subsection{Data processing} \label{subsec: preprocessing}

\subsubsection{Data cleaning} \label{subsubsec: cleaning}

Prior to analysis, it is necessary to clean the data of any records that might pollute or obscure our study. To do this, a set of procedures have been established to discard potential errors:
\begin{itemize}
    \item Duplicate rows and samples with missing buoy identification number are dropped from both the activity and the echo-sounder databases.
    \item Echo-sounder records corresponding to positions with less than 200m depth are removed, as the echo-sounder signal could be affected by the sea-floor. This filter also removes all acoustic records reported on land.
    \item Acoustic readings from buoys on board of vessels are removed by calculating the mean buoy velocity over a day and discarding rows where the buoy velocity exceeds 3 knots, following the same criterion as~\citet{Orue2019FromBuoys}.
\end{itemize}

\subsubsection{\textsc{Tun-AI} estimates} \label{subsubsec: tunai}

The biomass estimates provided by the echo-sounder may present variations when compared to real tuna tonnage under the dFAD ~\citep{Lopez2016ADevices,Escalle2019ReportData,Orue2019FromBuoys}. This could be due to multiple causes, including the influence of oceanographic conditions or the diverse species composition under the dFAD. To mitigate this issue, we estimate tuna biomass using \textsc{Tun-AI}~\citep{Precioso2022Tun-AI:Data} which has proven to be more accurate than simply considering the raw acoustic signal provided by the echo-sounder. \textsc{Tun-AI}, based on a Gradient Boosting (GB) algorithm~\citep{Friedman2001GreedyMachine} and trained using set and deployment events from the FAD logbook, uses information from the acoustic records, buoy location, and oceanographic variables to estimate the tuna biomass under dFADs. This pipeline includes:
\begin{enumerate}
    \item a binary classification model trained to estimate whether the tuna biomass under a dFAD is higher or lower than \tons{10}. This model has attained an $\fscore$ of 0.925.
    \item a regression model trained to give a direct estimate of the quantity of tuna biomass under a dFAD. This model has an error (MAE) of \tons{21.6} and a relative error (SMAPE) of $29.5\%$ when evaluated over sets.
\end{enumerate} 
Both models require a 72-hour echo-sounder window, containing one acoustic record per hour. \textsc{Tun-AI} also includes a 3-class classification model that will not be used in this study. For a detailed explanation of \textsc{Tun-AI}, we refer the reader to the original paper \citep{Precioso2022Tun-AI:Data}. One of the novelties of our analysis with respect to previous studies is the fact that using a regression model allows us to examine both the aggregation and disaggregation processes to dFADs, which would not be possible with a binary classification model.

\textsc{Tun-AI} models can provide hourly biomass estimates for each buoy, but this frequency is not adequate for our study due to the noise generated by the day-night oscillations in tuna biomass~\citep{Escalle2019ReportData}. To circumvent this problem, we generate daily biomass estimates for each buoy, producing \num{3873531} outputs in total, after the cleaning process described in Section~\ref{subsubsec: cleaning} is carried out. 

\subsubsection{Generating virgin segments}\label{subsubsec: splits}

To avoid the effects of potential human interactions when studying tuna aggregation dynamics under dFADs, the time series of each buoy was broken into smaller segments in which such processes were not altered by any external action, which we call \textit{virgin segments}.

To generate the virgin segments for any particular echo-sounder buoy, we first merge the \textsc{Tun-AI} estimates (for both the binary classification and the regression models) with the activity database, using the buoy identification number as primary key. Of the activities recorded in the FAD logbook, only deployments, sets, retrievals at sea, recoveries at port and losses were considered to be ``segment-generating'', that is, they could directly affect the echo-sounder readings and the biomass dynamics under the dFAD. Visits and modifications were assumed to have no effect on aggregated tuna biomass or on the echo-sounder readings, so they were not considered in this study. Lastly, a period of more than 24 hours with no information reported by the buoy would also generate a virgin segment, as this could indicate that the buoy was switched off or otherwise inoperable. 

\begin{figure}[ht] 
  \begin{subfigure}[t]{0.49\linewidth}
    \centering
    \includegraphics[width=\linewidth]{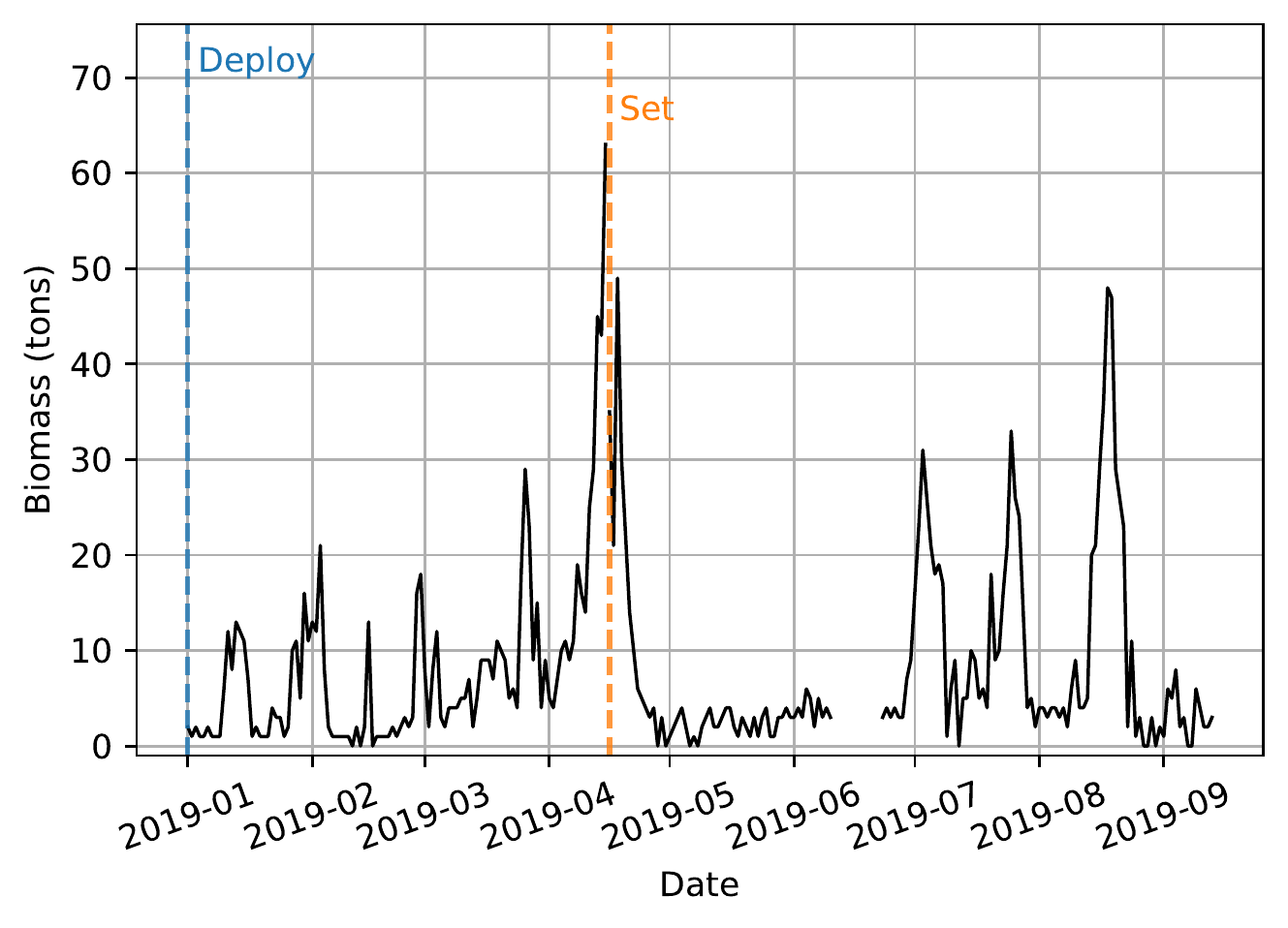} 
    \caption{A sample buoy's biomass estimates over time, together with registered activities from the FAD logbook depicted as dashed lines.} 
    \label{fig:tunai_sample1} 
  \end{subfigure}
  \quad
  \begin{subfigure}[t]{0.49\linewidth}
    \centering
    \includegraphics[width=\linewidth]{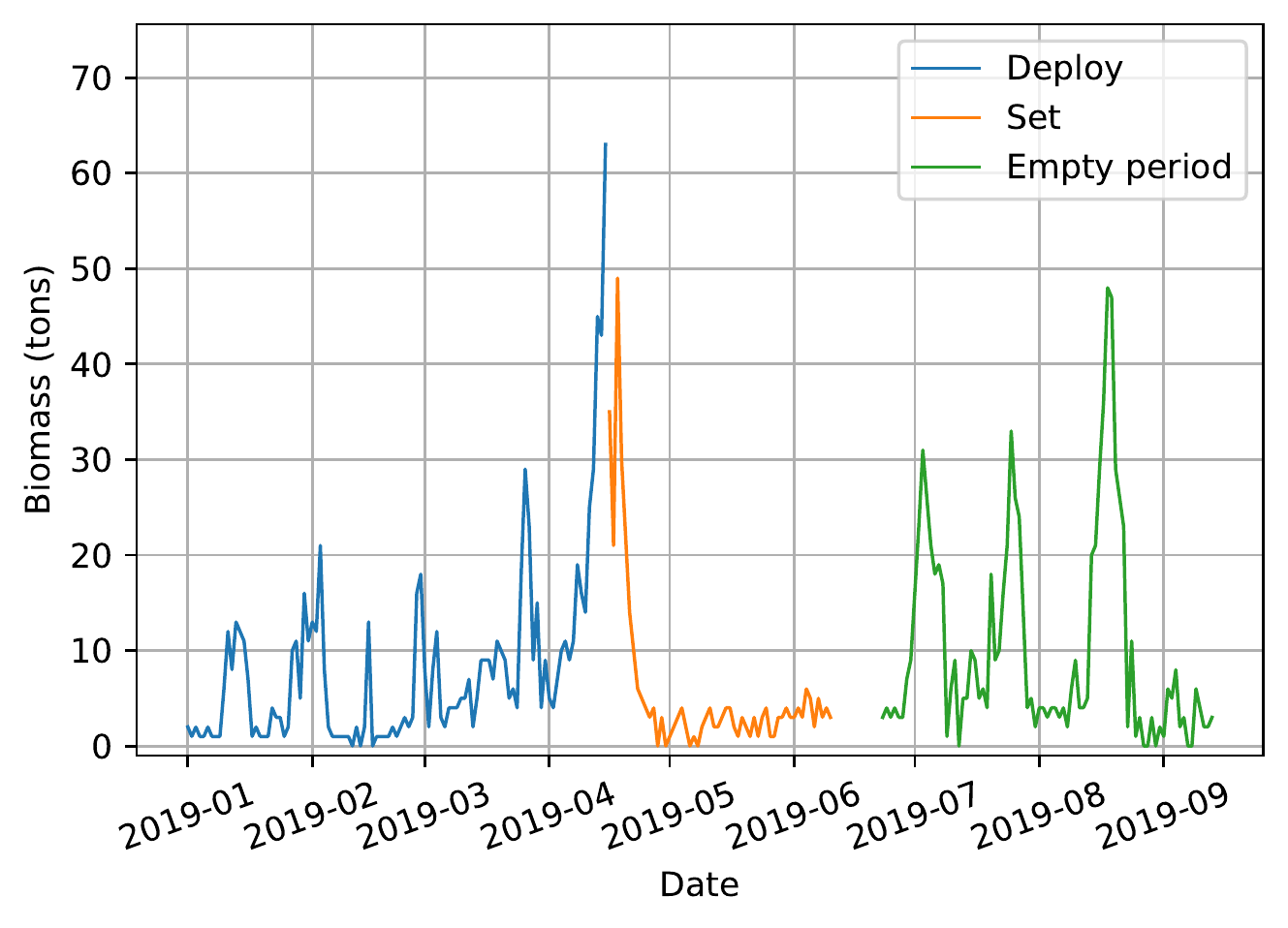} 
    \caption{Generation of the virgin segments from the activities registered in the FAD logbook, and from a period with no buoy records. Each colour represents the resulting individual virgin segments.} 
    \label{fig:tunai_sample2} 
  \end{subfigure} 
  \caption{Illustration of the process of generating virgin segments from the biomass estimates and registered FAD logbook activities on a sample echo-sounder buoy over time.}
  \label{fig:tunai_splits} 
\end{figure}

We only considered segments longer than 72h, as that is the minimum length of the window that \textsc{Tun-AI} needs to estimate biomass. We also omitted segments where \textsc{Tun-AI} failed to output an estimate for more than $80\%$ of the total segment length. This may happen for very short segments (not discarded previously because they are longer than 72h) or if the oceanographic data are not available (for example, due to issues on the CMEMS platform or with data resolution). Otherwise, missing values from \textsc{Tun-AI} were interpolated, in the case of the regression model, or propagated based on the last valid estimate, in the case of the binary classification model. Finally, after the pre-processing outlined in Section~\ref{subsubsec: cleaning}, and the steps described here, a total of \num{43334} virgin segments were generated. The process of generating the virgin segments is illustrated in Figure~\ref{fig:tunai_splits}.

\subsubsection{Smoothing the signal}\label{subsubsec: smoothing}

The output of both the binary and regression models of \textsc{Tun-AI} are more representative of real tuna biomass than the raw estimates provided by the buoy. However, some noise is still present in the data, likely due to the small-scale changes in tuna aggregations or to the influence of other fish species around the dFAD. Since the aim of the current study is to identify general trends in the tuna aggregation processes, we have smoothed the resulting series to capture general trends while discarding small oscillations. 

\begin{figure}[htbp]
    \centering
    \includegraphics[width=\linewidth]{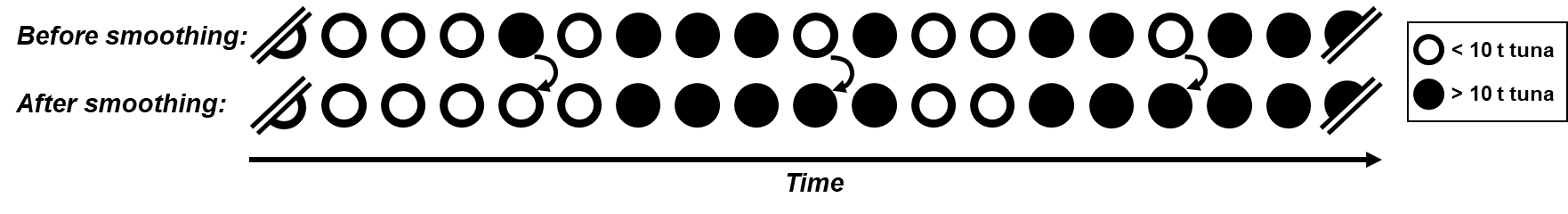}
    \caption{Schematic representation of the smoothing procedure for the binary series: isolated estimates are modified according to neighboring values.}
    \label{fig:binary_smoothing}
\end{figure}

For the binary series, isolated estimates of one class or another are smoothed according to the values recorded for the previous day (Figure~\ref{fig:binary_smoothing}).  Altogether, $2.7\%$ of the total binary data were modified by this smoothing procedure. 

\begin{figure}[htbp]
    \centering
    \includegraphics[width=\linewidth]{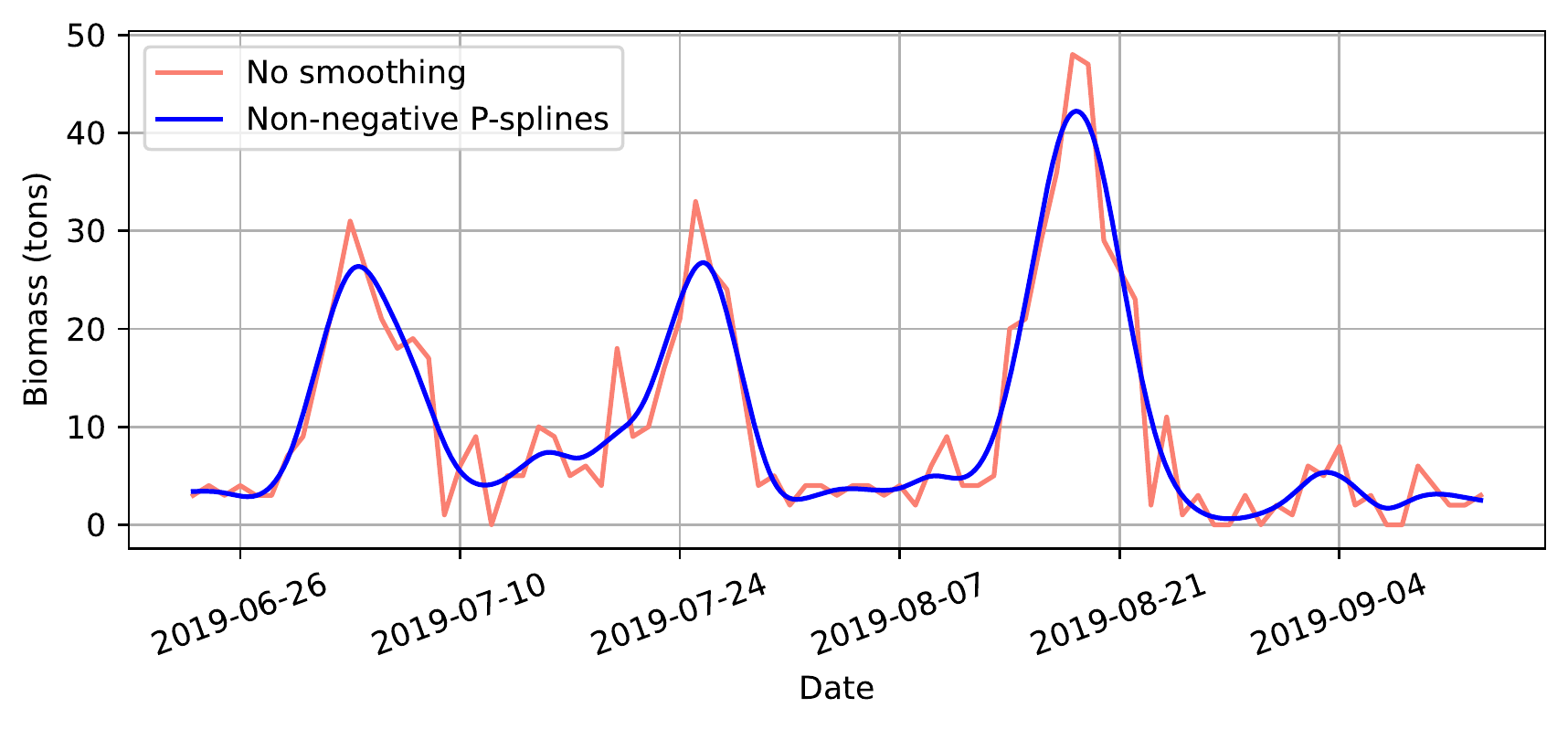}
    \caption{Illustration of the smoothing process for the regression series: Biomass estimates generated from the \textsc{Tun-AI} regression model (in pink) are smoothed to show general trends with the non-negative $P-$splines approach (in blue). The smoothed curve is less influenced by noise in the original data to better represent general trends while providing coherent estimates with the non-negative requirement.}
    \label{fig:buoy_smoothings}
\end{figure}

In the regression model, we applied a constrained $P-$splines approach developed in~\citet{Navarro-Garcia2022OnApproach}, which captures the trend of the data without overestimating the signal while forcing the response to be non-negative (as the nature of the data requires). To smooth the series following this methodology, the open source Python package \texttt{cpsplines} is used \citep{Navarro-Garcia2021Cpsplines}. Figure~\ref{fig:buoy_smoothings} shows the rightmost virgin segment in Figure~\ref{fig:tunai_sample2} together with its smoothed version. 

\subsubsection{Tuna dynamics characterization}\label{subsec: tuna_dynamics}

\begin{figure}[htbp]
    \centering
    \includegraphics[width=\linewidth]{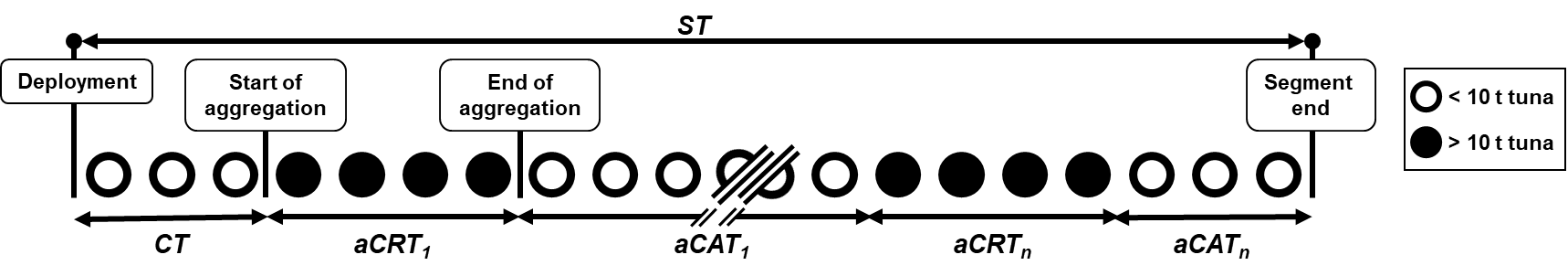}
    \caption{Schematic illustration of the results obtained by the \textsc{Tun-AI} binary model for a given dFAD, and the calculation of aggregation metrics based on virgin segments starting with a deployment. ST: Soak Time; CT: Colonization Time; aCRT: Aggregation's Continuous Residence Time; aCAT: Aggregation's Continuous Absence Time}
    \label{fig:soak_time}
\end{figure}

To characterize the temporal patterns of the tuna school's aggregation to newly deployed dFADs, we estimate a number of metrics using the binary classification results and virgin segments beginning with a deployment (\num{7368} segments):

\begin{itemize}
    \item Soak time (ST): reflects the amount of time a given dFAD has been drifting at sea. Thus, it is calculated here as the time elapsed between the initial deployment of the dFAD to the end of the virgin segment (Figure~\ref{fig:soak_time}).
    \item Colonization time (CT): captures the time between the initial deployment of the dFAD and the first detection of tuna~\citep{Orue2019AggregationSpecies}. Here, we estimate it as the time between the initial deployment of the dFAD and the first day where the binary model of \textsc{Tun-AI} outputs a positive prediction, i.e., tuna biomass is greater than \tons{10} (Figure~\ref{fig:soak_time}). 
    \item Aggregration’s Continuous Residence Time (aCRT): first defined as CRT by~\citet{Ohta2004PeriodicStations} for individually tagged tunas at dFADs, and adapted here to consider the entire aggregation, aCRT reflects how long a tuna aggregation is continuously detected by the echo-sounder buoy on a given dFAD without day-scale ($>24$h) absences. That is, aCRT is calculated as the number of days where \textsc{Tun-AI} has continuously estimated tuna biomass greater than \tons{10} (Figure~\ref{fig:soak_time}). 
    \item Aggregration’s Continuous Absence Time (aCAT): adapted here to consider the entire aggregation, this metric also draws from~\citet{Ohta2004PeriodicStations}. In a similar way, aCAT reflects how long the tuna is continuously absent from a given dFAD without day-scale ($>24$h) presences, and is calculated here as the number of days where \textsc{Tun-AI} has continuously estimated tuna biomass lower or equal than \tons{10} (Figure~\ref{fig:soak_time}).
    \item Occupancy Rate (OR): this variable is defined as the proportion of time that the tuna school remains at the dFAD after it has been colonized, and it can be estimated by means of the previous metrics.
    \item Percentage of dFADs that are never colonized: proportion of dFADs where the presence of tuna has never been observed. This is useful to contextualize the colonization time statistics.
\end{itemize}

\begin{figure}[htbp]
    \centering
    \includegraphics[width=\linewidth]{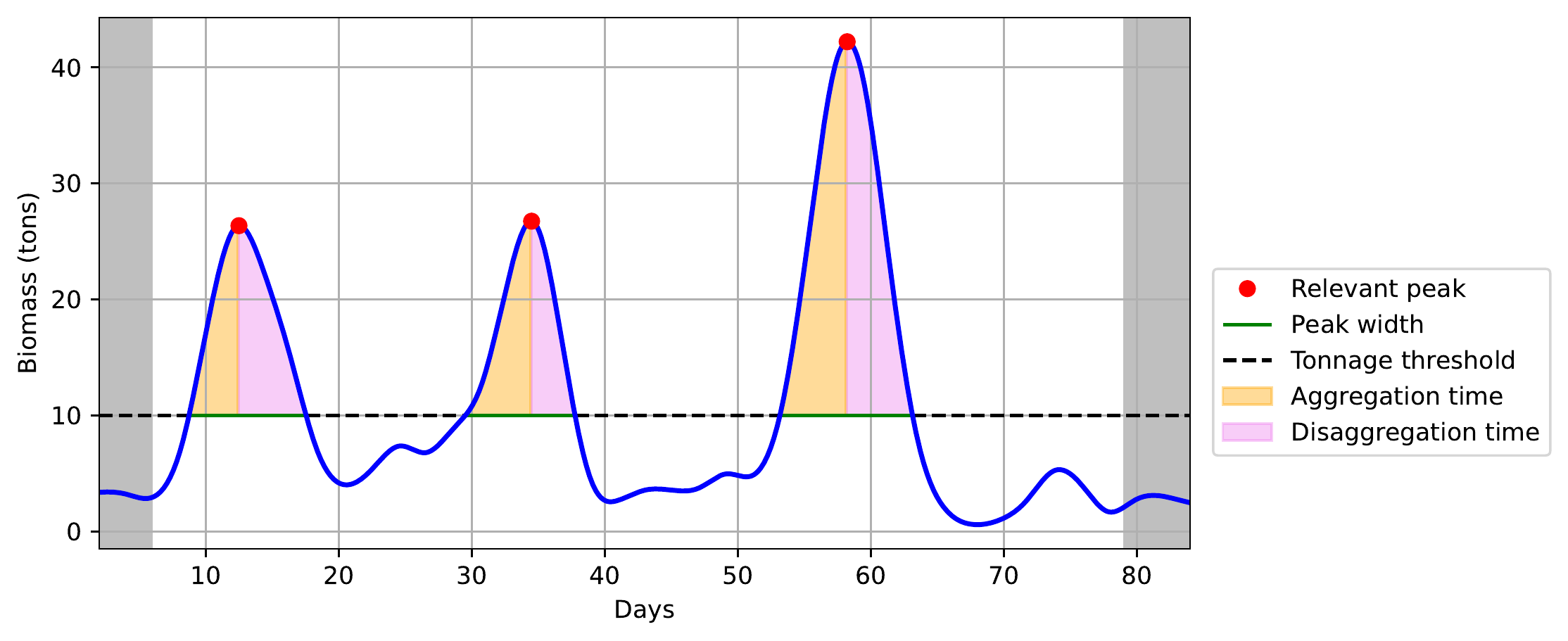}
    \caption{Schematic representation of the calculation of Aggregation Time (AT) and Disaggregation Time (DT) from the smoothed biomass estimates delivered by the \textsc{Tun-AI} regression model for a sample virgin segment. The shaded areas represent the days where no peaks are considered, and the dashed line represents the \tons{10} threshold.}
    \label{fig:peaks}
\end{figure}

Given that the current study also draws from estimates of the total amount of tuna under the dFAD, the processes of both aggregation and disaggregation can be examined. Therefore, we define two novel metrics: Aggregation Time (AT) and Disaggregation Time (DT). To calculate these, we consider the daily tuna biomass estimates provided by the \textsc{Tun-AI} regression model after smoothing (see Section~\ref{subsubsec: smoothing}). Using these data, we identified the moments where tuna biomass reaches a local maximum above 10t, since this is the amount of tuna we consider to be a significant aggregation. This was achieved using a modified version of \texttt{scipy.signal.find\_peaks} \citep{Virtanen2020SciPyPython}, and these peaks were determined by simple comparison of neighboring values of tuna biomass estimates. Any peaks found within the first or last 5 days of the virgin segment were discarded, as an extra precaution to avoid the effects of any human activity on the biomass estimates. This resulted in all virgin segments lasting less than 10 days to be discarded, so a total of \num{23326} virgin segments were considered. The final number of peaks is \num{71644}. For each peak, AT was then calculated as the time elapsed between the first biomass estimate larger than \tons{10}, to the day maximum biomass was reached. Likewise, DT was calculated as the time between the maximum biomass, to the next biomass estimate under \tons{10}. This process is illustrated in Figure~\ref{fig:peaks}. 

To better examine whether any of the previously mentioned metrics varied significantly across oceans, Kruskal-Wallis tests were carried out and followed by Dunn tests to confirm pairwise differences. Likewise, aCRT and aCAT, as well as AT and DT, were compared using Mann-Whitney tests.

\section{Results}\label{sec: results}

\subsection{General aggregation metrics}\label{subsec:results_binary}

\begin{figure}[htbp]
    \centering
    \begin{subfigure}[b]{0.49\textwidth}
        \centering
        \includegraphics[width=\textwidth]{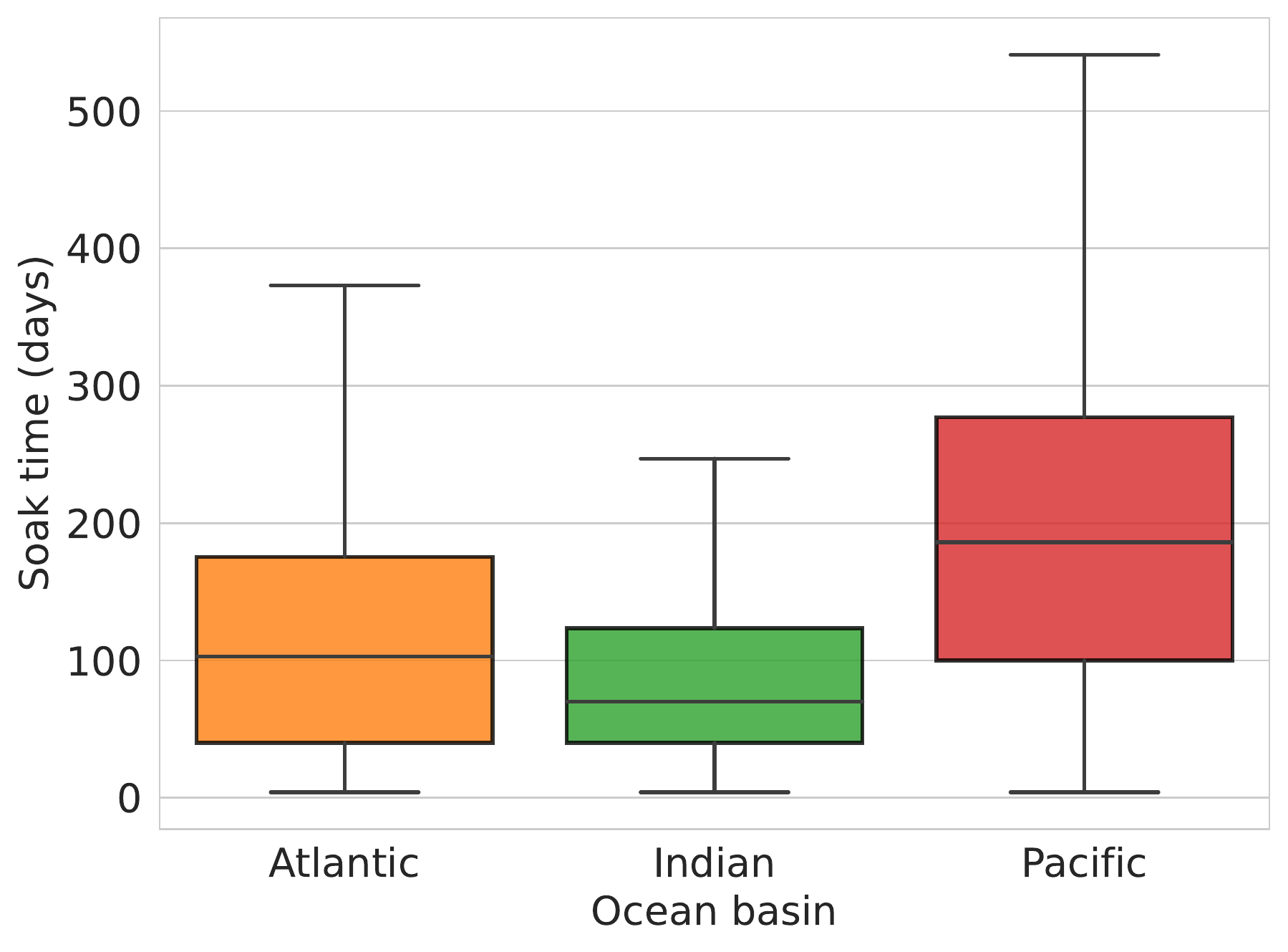}
        \caption{Soak time (ST)}    
        \label{fig:boxplot_st}
    \end{subfigure}
    \hfill
    \begin{subfigure}[b]{0.49\textwidth}  
        \centering 
        \includegraphics[width=\textwidth]{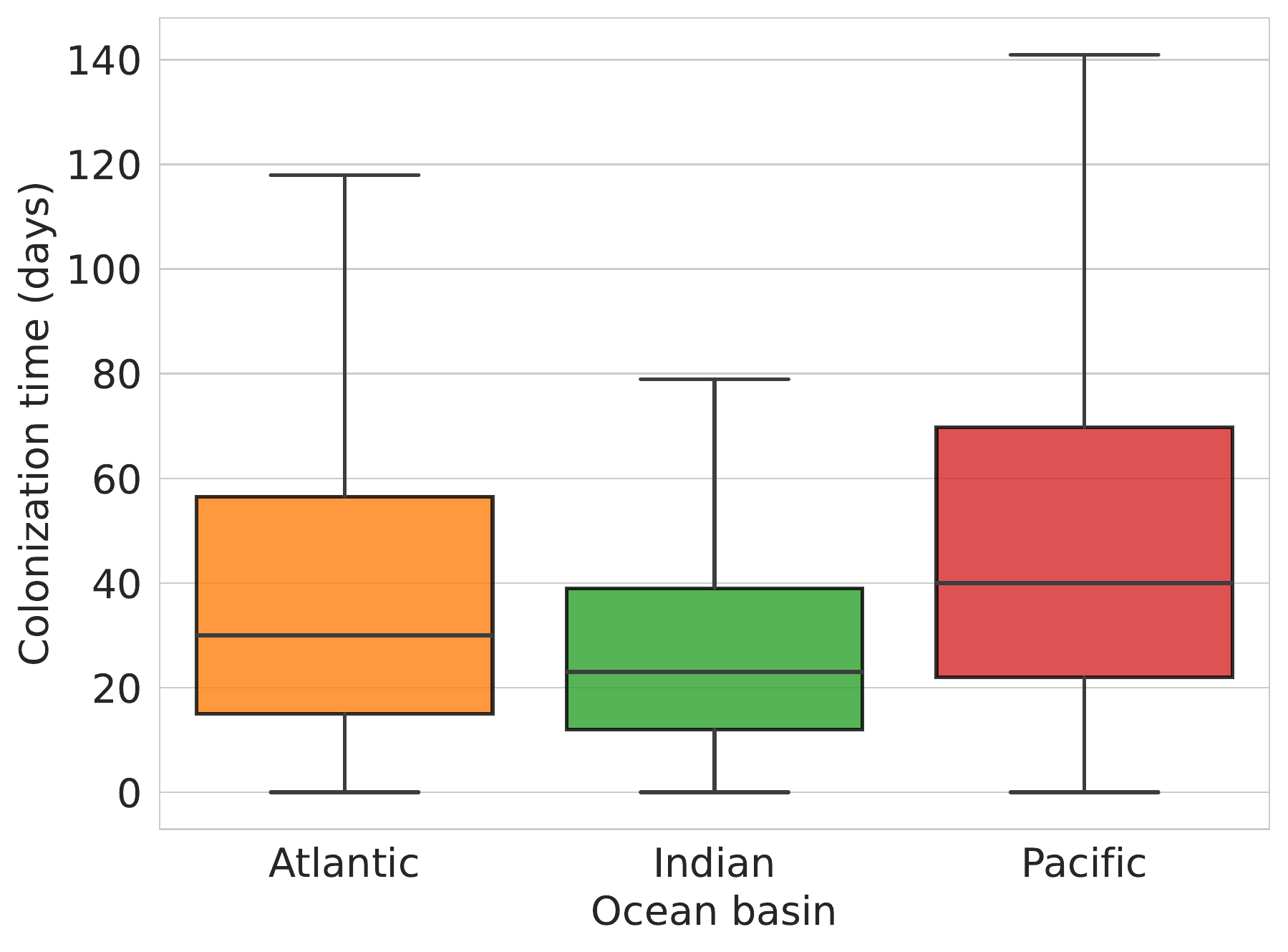}
        \caption{Colonization time (CT)}    
        \label{fig:boxplot_ct}
    \end{subfigure}
    \vskip\baselineskip
    \begin{subfigure}[b]{0.49\textwidth}   
        \centering 
        \includegraphics[width=\textwidth]{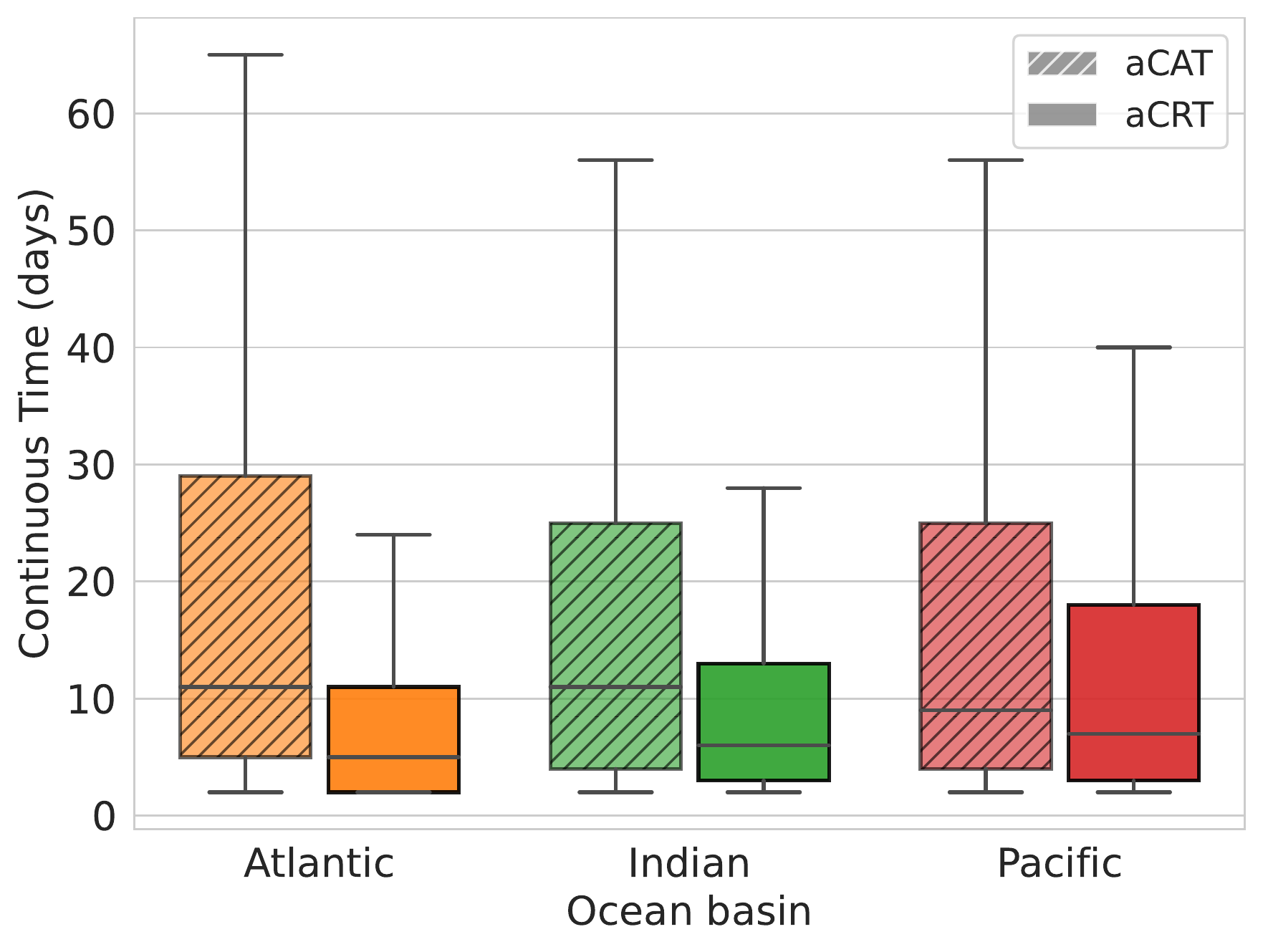}
        \caption{Aggregation continuous residence and absence time (aCRT and aCAT)}    
        \label{fig:boxplot_continuous}
    \end{subfigure}
    \hfill
    \begin{subfigure}[b]{0.49\textwidth}   
        \centering 
        \includegraphics[width=\textwidth]{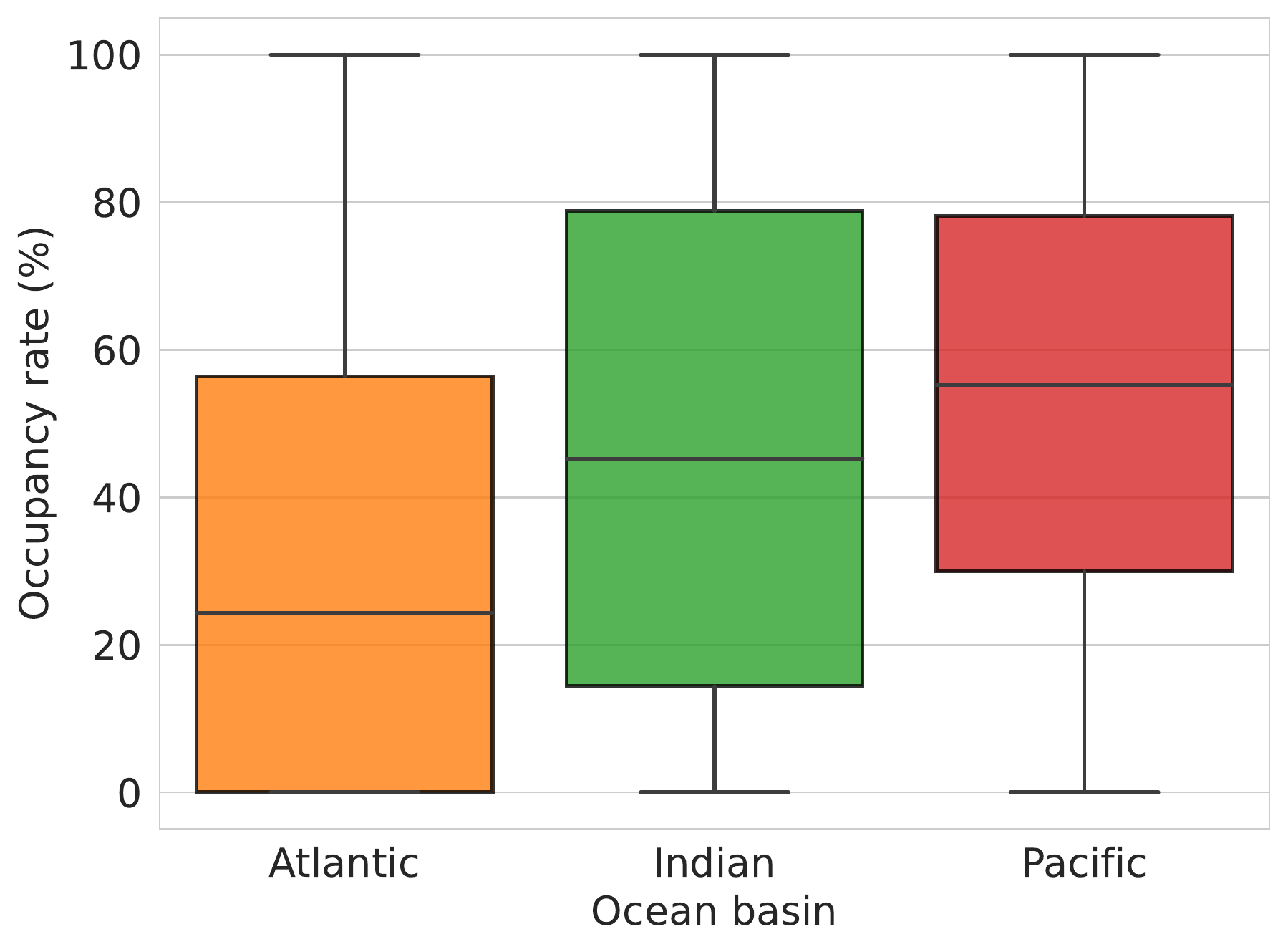}
        \caption{Occupancy rate (OR)} 
        \label{fig:boxplot_or}
    \end{subfigure}
    \caption{Box plots of the variables estimated from the binary model. Outliers where removed from the figure using the interquartile rule of thumb with parameter 1.5.}
    \label{fig:boxplots}
\end{figure}

Regarding the variables estimated from the binary \textsc{Tun-AI} biomass estimates and using newly deployed dFADs, a summary of statistical metrics, itemized by the ocean basin, is displayed in Table~\ref{tab:binary_metrics}, and their distributions are shown by means of box plots in Figure~\ref{fig:boxplots}.

\begin{table}[htbp]
\centering
\sisetup{round-precision=0}
\small
\begin{tabular}{llS[table-format=5]
S[table-format=3]
S[table-format=3]
S[table-format=3]
S[table-format=3]}
\toprule
Metric & Ocean & {Count} & {Mean} & {SD} & {Median} & {IQR} \\
\midrule
ST (days) & Atlantic & 1015 & 124 & 101 & 103 & 136 \\
& Indian & 1591 & 92 & 73 & 70 & 84\\
& Pacific & 4762 & 202 & 129 & 186 & 177 \\
[0.5em]
CT (days) & Atlantic & 1015 & 44 & 43 & 30 & 42 \\
& Indian & 1591 & 29 & 25 & 23 & 27 \\
& Pacific & 4762 & 51 & 42 & 40 & 48 \\
[0.5em]
aCRT (days) & Atlantic & 3201 & 10 & 16 & 5 & 9 \\
& Indian & 4389 & 11 & 16 & 6 & 10 \\
& Pacific & 24408 & 17 &25 & 7 & 15 \\
[0.5em]
aCAT (days) & Atlantic & 3875 & 24 & 33 & 11 & 24 \\
& Indian & 5088 & 19 & 23 & 11 & 21 \\
& Pacific & 26552 & 21 & 30 & 9 & 21 \\
[0.5em]
OR (\%) & Atlantic & 1015 & 33 & 32 & 24 & 56 \\
& Indian & 1591 & 48 & 35 & 45 & 64 \\
& Pacific & 4762 & 53 & 31 & 55 & 48 \\
\bottomrule
\end{tabular}
\caption{Summary statistics, per ocean, for tuna aggregation metrics calculated from virgin segments starting with a deployment. SD refers to the standard deviation and IQR denotes the interquartile range.}
\label{tab:binary_metrics}
\end{table}

In terms of the ST and the CT, both show similar patterns between oceans (Figures~\ref{fig:boxplot_st},~\ref{fig:boxplot_ct}): the longest CT and ST are reported in the Pacific Ocean, and the shortest in the Indian Ocean, while the Atlantic Ocean showed results somewhere in between the previous two. In fact, median ST for the Pacific Ocean more than doubled that of the Indian Ocean, while CT nearly doubled it. Concerning the variability of these variables, the lowest standard deviation occurs in the Indian Ocean, while results were more variable for the other two oceans. The proportion of dFADs that were not colonized throughout their soak time also presented considerable variations (27\% in the Atlantic, 16\% in the Indian and 11\% in the Pacific).

Regarding the aCRT, aCAT, and OR, consistent patterns are again visible across oceans, although trends are different for ST and CT (Figure~\ref{fig:boxplot_continuous}, ~\ref{fig:boxplot_or}). In this case, the Indian Ocean showed values for aCRT, aCAT and OR that fell between those of the Atlantic and Pacific Oceans. Concerning the aCRT, the Atlantic Ocean showed the shortest times, and the Pacific Ocean showed the longest, while the opposite was true for aCAT (Figure~\ref{fig:boxplot_continuous}). Hypothesis tests showed significant differences for both aCRT and aCAT among oceans (Kruskal-Wallis test, $p < 0.01$) and these differences were confirmed in the pairwise comparisons between oceans (Dunn test, $p < 0.01$). Median values for both aCRT and aCAT across oceans were generally similar, ranging from 5 to 7 days, or 9 to 11 days, respectively (Table~\ref{tab:binary_metrics}). Overall, global aCRT was significantly lower than the global aCAT (Mann-Whitney test, $p \gg 0.01$), and variability was also consistently higher for aCAT than for aCRT. Lastly, OR was globally around 50\%, with the lowest median OR is registered in the Atlantic Ocean at 24\% (Table~\ref{tab:binary_metrics}). 

\subsection{Aggregation and disaggregation times}\label{subsec:results_continuous}

\begin{table}[htbp]
\footnotesize
\centering
\begin{tabular}{
l
S[table-format=5]
S[table-format=3.1,round-precision=1]
S[table-format=3.1,round-precision=1]
S[table-format=2.1,round-precision=1]
S[table-format=3.1,round-precision=1]
S[table-format=3.1,round-precision=1]
S[table-format=3.1,round-precision=1]
S[table-format=2.1,round-precision=1]
S[table-format=3.1,round-precision=1]}
\toprule
& & \multicolumn{4}{c}{Aggregation time (days)} & \multicolumn{4}{c}{Disaggregation time (days)} \\
\cmidrule(lr){3-6}\cmidrule(lr){7-10}
Ocean & {Count} & {Mean} & {SD} & {Median} & {IQR} & {Mean} & {SD} & {Median} & {IQR} \\ 
\midrule
Atlantic & 19581 & 7.60 & 13.51 & 3.02 & 5.64 & 6.36 & 10.47 & 3.00 & 4.67 \\
Indian   & 26806 & 8.24 & 13.65 & 3.63 & 6.85 & 7.26 & 12.48 & 3.44 & 5.42 \\ 
Pacific  & 25257 & 15.45 & 25.04 & 5.89 & 13.99 & 14.63 & 25.99 & 5.49 & 11.03 \\
\bottomrule
\end{tabular}
\caption{Summary of tuna aggregations metrics for the continuous model and decoupled by ocean basin.}
\label{tab:number_peaks}
\end{table}

\begin{figure}[htbp]
    \centering
    \includegraphics[width=\linewidth]{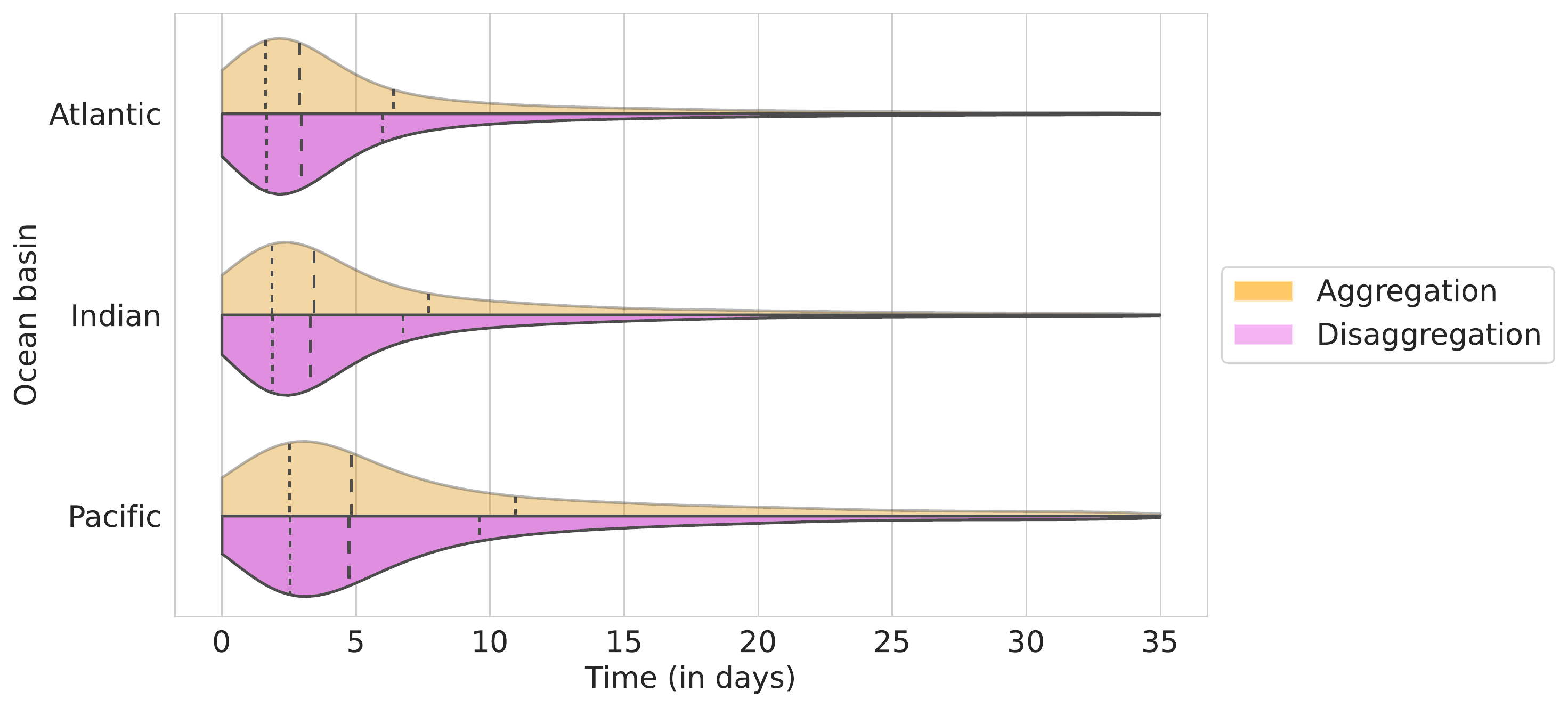}
    \caption{Violin plots of the aggregation and disaggregation time distributions itemized by ocean basin. The vertical dashed lines depicted the position of the quartiles of the distributions.
}
    \label{fig:agg_disagg}
\end{figure}

Using the \textsc{Tun-AI} regression model, we were able to examine tuna aggregation dynamics around dFADs with more detail, estimating both AT and DT. In general, AT and DT showed similar patterns across oceans, with the shortest median AT and DT being registered for the Indian Ocean, and the longest for the Pacific Ocean (Table~\ref{tab:number_peaks}, Figure~\ref{fig:agg_disagg}). Globally, DT was not significantly longer than AT  (Mann-Whitney test, $p \gg 0.01$). In fact, it is worth noting that the first quartile for both AT and DT was generally similar, while more variation was seen for the third quartile, with AT generally longer than DT (Figure~\ref{fig:agg_disagg}). Significance tests found differences for both AT and DT among oceans (Kruskal-Wallis test, $p < 0.01$) and in the pairwise comparisons between oceans (Dunn test, $p < 0.01$). Lastly, the distributions for AT and DT were positively skewed (i.e., the mean was greater than the median) regardless of the ocean where the dFAD was deployed.

\section{Discussion}\label{sec: discussion} 

By using the data acquired by echo-sounder buoys attached to dFADs, over the course of several years and across all oceans, the current study aimed to capture the general trends in tuna aggregation dynamics at a global scale. This was achieved by means of a powerful Machine Learning pipeline, \textsc{Tun-AI}~\citep{Precioso2022Tun-AI:Data}, which processes echo-sounder information to deliver estimates of tuna tonnage under each dFAD either as a binary output ($<\tons{10}$ or $\geq\tons{10}$), or as a direct estimate of biomass. To the best of our knowledge, this is the first analysis to examine typical metrics of tuna aggregation (ST, CT, aCAT, aCRT and OR) across all oceans and in such detail, providing insight into both the processes of aggregation and disaggregation in tuna’s colonization around dFADs. 

Using a binary model, and applying a similar approach with echo-sounder buoys from a different manufacturer, ~\citet{Baidai2020TunaBuoys} quantified several metrics related to tuna aggregation's around dFADs in the Atlantic and Indian Oceans. In terms of ST, their estimates are considerably shorter than ours for the same oceans (median values of 44 and 43 days in the Atlantic and Indian Oceans, respectively), likely due to slight differences in definition. While~\citet{Baidai2020TunaBuoys} define ST as “the number of days between the deployment of a DFAD equipped with a buoy and the first reported operation on it”, our definition captures the length of the entire virgin segment, which would likely be longer for buoys where no activities other than deployment were registered, which made up 27\% of our dataset. In the Pacific Ocean, \citet{Escalle2021ReportMeetings} reported mean drift times of 118 days for dFADs included in the Parties to the Nauru Agreement’s (PNA) FAD tracking trial program, which is shorter than the median 202 days ST in our results. However, she highlights that due to data sharing constraints, it is likely that these times are underestimated, as information outside of the PNA’s Exclusive Economic Zones was not analyzed. Fishermen have mentioned that the average lifespan of an artificial dFAD is about 5–12 months~\citep{Lopez2017EnvironmentalBuoys}, in line with the values registered here.

In terms of CT, it appears that there is no general consensus among fishing masters. \citet{Moreno2007FishDFADs} interviewed fishing masters from the Indian Ocean, of which about one third considered that it usually takes a minimum of 1 month for a school to aggregate to a dFAD. Indeed, although there is considerable variation in the CTs registered in our study for dFADs across all oceans, median values are around 20–40 days, in line with the observations of these fishing masters. However, about 45\% of interviewed fishing masters believed that tuna’s colonization of a dFAD was not dependent on time~\citep{Moreno2007FishDFADs}, an observation which was also reflected in~\citet{Lopez2017EnvironmentalBuoys}, where tuna abundance at dFADs was not positively correlated with ST, and evidenced in the large variability for CT in our data.

Although extensive literature has been dedicated to examining the time spent by tunas both near and away from floating objects, most work has been conducted on individually tagged tunas at a limited number of study sites~\citep{Govinden2021BehaviorTelemetry, Schaefer2005BehaviorPacific, Tolotti2020AssociationOcean, Dagorn2007BehaviorFADs, Ohta2004PeriodicStations,Chiang2021Fine-scaleTaiwan,Matsumoto2016ComparisonOcean,Matsumoto2014BehaviorOcean,Rodriguez-Tress2017AssociativeMauritius}. These methods provide high levels of detail, but also may not be representative of overarching trends across all dFADs, nor even the general patterns of an entire school of tuna. For example,~\citet{Robert2012Size-dependentFADs} found size dependent differences in the time yellowfin spent around an anchored FAD, with smaller individuals ($<50$cm fork length) spending about four times as much time around the FAD than larger individuals. Similarly, differences in the CRT of skipjack, yellowfin and bigeye tuna have been observed across oceans~\citep{Govinden2021BehaviorTelemetry, Schaefer2005BehaviorPacific, Tolotti2020AssociationOcean}. 

Even though the massive data that is available from echo-sounder buoys attached to dFADs may not provide such highly detailed information, it does show potential for identifying general trends on how entire aggregations of tuna behave. As in our study, \citet{Diallo2019TowardsM3I+} used echo-sounder data from two dFAD buoy models from a different manufacturer to estimate aCRT and aCAT in the Indian Ocean. Both aCRT and aCAT were shorter than ours, and significant differences were found between buoy models ($6–8$ and $8–9$ days, respectively, depending on buoy model). This is an important factor to consider when comparing the results of different studies using echo-sounder buoys. \citet{Diallo2019TowardsM3I+} conclude that the higher sensitivity of the newer model could be driving the differences in aCRT and aCAT, so it stands to reason that buoys from different manufacturers would also register biomass differently. For example, the use of different frequency echo-sounders likely impacts the biomass estimates provided by different buoy brands~\citep{Lopez2014EvolutionOceans, Moreno2019TowardsDevices}. Indeed, fishing masters perceive differences in the biomass readings of different manufacturers~\citep{Lopez2014EvolutionOceans}, so these differences should be handled with care. Fishing technology evolves quickly, and it is important for researchers to be in line with manufacturers when drawing conclusions from technology derived data. Nonetheless, even between buoy brands, and across oceans, aCRT and aCATs are generally less than 10 days~\citep{Baidai2020TunaBuoys,Diallo2019TowardsM3I+}. This is in accordance with the median and average values of CRT found by most other authors when examining individual tunas around dFADs (see~\citet{Baidai2020TunaBuoys} and references therein).

In a wider context, one of the main concerns around dFAD use has been centered on the possibility that dFADs could constitute an ecological trap, whereby tuna remain associated to the dFAD even as it drifts into areas that are not favorable for the tuna’s growth and development~\citep{Marsac2000DriftingTrap, Hallier2008DriftingSpecies}. While other authors have reviewed available literature and concluded that there was not sufficient evidence to support or reject this hypothesis~\citep{dagorn2012IsEcosystems}, further research has been called for. One of the novel aspects of the current study was the application of a regression model to the echo-sounder buoy data, which allowed for direct estimates of tuna biomass aggregated to the dFAD~\citep{Precioso2022Tun-AI:Data}, and the calculation of two derived metrics: AT and DT, which could provide further insight into the nature of tuna's association to dFADs. Given that one of the premises for an ecological trap to be happening is that the tuna's association to the dFAD is ``fast, strong, and long-lasting'' \citep{Marsac2000DriftingTrap}, it would be reasonable to expect DT to be longer than AT if the dFAD were indeed ``trapping'' the tuna. However, our results showed that this was not the case, since DT was not significantly longer than AT. In fact median AT and DT values generally did not show differences longer than a day and, where differences were present, such as in the third quartile, the time it took the aggregation of tuna to depart was actually shorter than it took for the aggregation to form in the first place. Although these results should be explored further, at a global scale there does not appear to be evidence of an ecological trap.  

While the current study has focused on the temporal patterns of tuna aggregation to dFADs, future research could focus on the spatial dynamics at play. Fishing masters interviewed by~\citet{Moreno2007FishDFADs} stated that the departure of tuna schools from dFADs was often related to changes in currents or FAD drift trajectory, and that changes in the surrounding environment, such as temperature, could also cause tuna to leave the dFAD. Whether this is indeed occurring could be tested, by examining the oceanographic context around dFADs during aggregation and disaggregation processes. Likewise, testing whether local dFAD density has an effect on CT, aCRT or aCAT could serve to assess optimal dFAD usage for purse-seine fleets. Though there are inherent challenges in using data provided by echo-sounder buoys attached to dFADs, the present study is an excellent example of how this information, combined with Data Science techniques for filtering and processing, can provide a cost effective tool for shedding light on tuna behaviour and biology.

\section*{Acknowledgments}

This study has been conducted using E.U. Copernicus Marine Service Information. We also thank AGAC for providing the logbook data used in the analysis and the helpful comments about the manuscript. The authors would also like to thank Carlos Roa for rendering available the Satlink echosounder dataset. The research of DGU has been supported in part by the Spanish MICINN under grants PGC2018-096504-B-C33 and RTI2018-100754-B-I00,  the European Union under the 2014-2020 ERDF Operational Programme and the Department of Economy, Knowledge, Business and University of the Regional Government of Andalusia (project FEDER-UCA18-108393). The research of Manuel Navarro-García has been financed by the research project IND2020/TIC-17526 (Comunidad de Madrid). The research of Alberto Torres has been financed in part by a Torres Quevedo grant PTQ2019-010642 from Agencia Estatal de Investigaci\'on (Spain). The research of Daniel Precioso has been financed by an Industrial PhD grant from the University of Cádiz.

\bibliographystyle{apalike}
\bibliography{references}

\end{document}